# On Implementing Usual Values

Ronald R. Yager
Machine Intelligence Institute
Iona College
New Rochelle, N.Y. 10801

## Abstract

In many cases commonsense knowledge consists of knowledge of what is usual. In this paper we develop a system for reasoning with usual information. This system is based upon the fact that these pieces of commonsense information involve both a probabilistic aspect and a granular aspect. We implement this system with the aid of possibility-probability granules.

## Introduction

An ability to handle commonsense reasoning is a crucial need in the development of the artificial intelligence [1]. A number of variants of so-called non-monotonic logics have been introduced as aids in developing these commonsense reasoning mechanisms [2-4].

Recently L.A. Zadeh has suggested that a central role in a theory of commonsense must be played by a concept of "usuality". This concept reflects the fact that in many cases commonsense reasoning involves reasoning with usual values for variables. For example, the statement
"a cup of coffee costs fifty cents,"
is more precisely reflected by the statement
"a cup of coffee usually costs fifty cents."

In a number of recent presentations Zadeh [5-7] has suggested some properties which must be present in any theory of usuality. In this paper we introduce a formal mechanism for representing and manipulating usual values. This formalism allows for both logical and arithmetic manipulations of usual values. The formal structure used to represent these usual values are Possibility-Probability granules (Poss-Prob granules). These structures have been studied by Yager [8-10] and are based upon a combination of the linguistic variables introduced by L.A. Zadeh in his theory of approximate reasoning [11] and the evidential structures introduced by G. Shafer [12].

As we shall see, the idea of usuality implies some random or probabilistic phenomenon at play as well as some idea of granularity. The idea of granularity necessitates the use of set theoretic constructs in the form of possibility distributions which enable us to account for the fact that humans conceptualize in terms of gross concepts. The introduction of the probabilistic aspect also provides a departure from the approach taken by the non-monotonic logicians.

We further note that these usual values play a central role as default values in frames and other similar type structures [13-15].

Since much of the information in knowledge-based systems involves the use of commonsense knowledge. This theory of usuality will greatly impinge in this area.

## On Possibility-Probability Granules

In preparation for presenting our representation scheme for usual values we briefly review the idea of possibility-probability granules. For more details on these structures see [8-10].

339

Assume V is a variable which takes its value in the set X. Assume A is a fuzzy subset of X a <u>canonical statement</u> of knowledge about V is of the form <u>V is A</u>. A canonical statement is meant to be interpreted as a proposition indicating that the value of V lies in the subset A. These type of statements provide a restriction on the value of V. A canonical statement provides a formal way of representing many natural language statements. Consider the statement "Bill is young." This can be represented as V is A where V would indicate the attribute Bill's age and A would be the subset of ages constituting the user's definition of young. Canonical statements of this type provide a mechanism for associating values to attributes when there exists some uncertainty as to the value of the attribute. In particular a canonical statement associates a set of <u>possible</u> values, those elements in the set A, with the variable V. We denote the type of uncertainty associated with these statements as possibilistic uncertainty.

A more general data structure which subsumes these canonical statements is what we shall call a Possibility-Probability granule. Again assume V is a variable which takes its value in the set X. Let $A_k$, k = 1,2,........n, be a collection of fuzzy subsets on X. Again in this framework <u>V is $A_k$</u> is a canonical statement restricts the value of the variable V. However in this more general setting of Poss-Prob granules we allow for some probabilistic uncertainty as to which is the appropriate canonical statement restricting the value of V. In particular we associate with each statement <u>V is $A_k$</u> a probability $p_k$, which indicates the probability that V is $A_k$ is the appropriate proposition expressing our knowledge about V. We note that their exists no restriction on the relationship between the $A_k$'s, other than they be non-null fuzzy subsets of X, however the $p_k$'s must sum to one.

Care should be taken to understand that $p_k$ is not the probability that the value of an experiment on X results in an element in $A_k$ but more in the following vain. There exists some other space $Y = \{y_1, y_2, ......, y_n\}$ in which we perform a random experiment in which $p_k$ is the probability that $y_k$ is the outcome. If as a result of this experiment $y_k$ occurs then we say that <u>V is $A_k$</u> is the canonical statement restricting V.

In [10] Yager has shown that the structure captured by these Poss-Prob granules is similar to that of a Dempster-Shafer belief structure in which the $A_k$'s are the focal elements and the $p_k$'s are the weights. However, we note in this Poss-Prob framework the $A_k$'s can be fuzzy subsets. Because of the similarity with the Dempster-Shafer belief structure we shall denote the knowledge that a variable's value is controlled by a Poss-Prob granule as

$$V \text{ is } m,$$

where m is a basic probability assignment function (bpa) with focal elements $\{A_k\}$ and weights $m(A_k) = p_k$.

Two important concepts associated with Dempster-Shafer are the measures of plausibility and belief defined on subsets of X. Because of the allowance for fuzzy sets as focal elements we must provide more general definitions for these measures in the framework of Poss-Prob granules. For any subset B of X
$Pl(B) = \sum_k (Poss[B/A_k] * m(a_k))$ and $Bel(B) = \sum_k (Cert[B/A_k] * m(A_k))$, where $Poss[B/A_k] = Max_x [B(x) \wedge A_k(x)]$ and $Cert[B/A_k] = 1-Poss[B^-/A_k]$. In the above $B(x)$ and $A_k(x)$ are membership grades of x in the respective sets and $B^-$ is defined by $B^-(x) = 1-B(x)$.

In [10] Yager has shown that when the focal elements are restricted to be crisp subsets of X these definitions collapse to those of Shafer.

As is well established in the Dempster-Shafer setting these two measures provide bounds on the probabilities of events in the space X of

340

outcomes for V. In particular, if we denote Prob(B) ti indicate the
probability that the value for V lies in the set B then
$$Bel(B) \leq Prob(B) \leq Pl(B).$$
The use of Poss-Prob granules provides a very powerful mechanism for representing various different types of knowledge about the variable V in a unified structure. One particular type of knowledge we shall find useful in this paper is the case in which we know that Prob(B) is "at least $\alpha$." Formally this can be stated as $\alpha \leq Prob(B) \leq 1$. Thus in this case we require that Bel(B) = $\alpha$ and Pl(B) = 1. It can easily be shown that the least restrictive Poss-Prob granule which can represent this information is a bpa m on X such that m(B) = $\alpha$ and m(X) = 1-$\alpha$.

## On Usual Values and Their Representation

As we noted in a number of recent papers L.A. Zadeh has introduced the concept of usuality and discussed its central role in any theory of commonsense reasoning. In particular Zadeh argues that in many cases the types of knowledge which constitutes commonsense involves the knowledge about the _usual_ value of some variable. Furthermore we understand Zadeh to suggest that in many cases these usual values are vague and imprecise of the type best represented by a linguistic value and the associated ideas of a fuzzy subset and a possibility distribution. This imprecision is due to the granular nature of human conceptualization.

If V is a variable taking its value in the set X and A is a fuzzy subset of X representing the usual value of V, we could then say, "usually V is A" or equivalently U(V) is A, where U(V) denotes the usual value of V. Examples of the above of linguistic structure would be
   "usually basketball players are tall,"
   "usually birds fly"
   "usually Mary comes home at about 8 o'clock.
In many instances of natural language discourse we suppress the word usually and simply say, for example, "birds fly" rather than "usually birds fly." If one doesn't recognize this shorthand many difficulties follow.

The statement "usually Mary comes home at 8 o'clock," as formalized by Usually V is A embodies a number of different forms of uncertainty. We see that the statement U(V) is A implies some probabilistic phenomenon in our knowledge about the variable V. We shall now provide a formal framework for representing this type of knowledge.

According to Zadeh the statement Usually V is A should be interpreted as indicating that the probability that the event A occurs as the value for the variable V is "at least $\alpha$", where $\alpha$ is some number close to one. The usual the occurrence of A the closer $\alpha$ is to one.

As we have indicated in the previous section this type of information on the variable V can be represented as a Poss-Prob granule. In particular the knowledge that usually V is A can be represented as the Poss-Prob granule
   V is m,
where m is a bpa on X, the frame of V, such that m(A) = $\alpha$ and m(X) = 1-$\alpha$.

Thus we see that the effect of the statement usually V is A is to say that $\alpha$ portion of the time the value of V is determined by the proposition V is A and that for 1-$\alpha$ of the time V is unknown.

The form V is m shall constitute a canonical type of representation for usual information. In the next section we shall provide for the translation of various linguistic structures involving usual values into these structures.

Before preceding we note that a non-probabilistic assertion such as "John is about 30 years old" can be written in this formation as V is $m_1$

341

where V is the attribute John's age and $m_1$ is a bpa such that $m_1(B) = 1$ where B is "about thirty."

## Translation of Compound Statements

In this section we shall provide some procedures for translating compound linguistic statements involving usual values into formal structures in terms of Poss-Prob granules. Our purpose here is to put these complex linguistic statements into forms which enable us to use the sophisticated mechanisms available for combining these structures as necessary in the course of the reasoning process.

The approach here is based upon a generalization of the approach used in Zadeh's theory of approximate reasoning [11].

We first start with the representation of linguistic structures in which propositions involving linguistic variables are qualified by the modifier "usually."

Assume V is a variable taking values in the set X. We recall a statement of the form <u>V is A</u>, where A is a fuzzy subset of X, is called a <u>canonical proposition</u>. As we discussed in the previous section the effect of the qualification of this proposition by usually to Usually (V is A) is to transform this statement into a Poss-Prob granule of the form <u>V is $m^*$</u> where $m^*$ is a bpa on X such that $m(A) = \alpha$ and $m(X) = 1-\alpha$.

Assume $V_1$ and $V_2$ are two variables taking their values in the sets X and Y respectively. Consider the conditional statement
"if $V_1$ is A then $V_2$ is B,"
where A and B are fuzzy subsets of X and Y respectively. From the theory of approximate reasoning, this conditional statement translates into compound canonical propositions $(V_1, V_2)$ is H where H is a fuzzy subset of X × Y which can be defined by
$$H(x, y) = \text{Min}[1, 1-A(x) + B(y)].$$
Alternatively H can be defined as $H(x,y) = \text{Max}[1-A(x), B(y)]$.
Now consider the qualified version of this statement "usually if $V_1$ is A then $V_2$ is B." This can be seen to be equivalent to <u>usually $(V_1, V_2)$ is H</u> which can be represented as any usuality qualified canonical proposition as a Poss-Prob granule <u>$(V_1, V_2)$ is m</u> where m is a bpa on X × Y such that $m(H) = \alpha$ and $m(X \times Y) = 1-\alpha$.

Consider now the statement <u>usually ($V_1$ is A or $V_2$ is B)</u>. Since the statement <u>V is A or $V_2$ is B</u> translates into the compound canonical proposition $(V_1, V_2)$ is $H^*$ where H is a fuzzy subset of X × Y such that $H^*(x, y) = \text{Max}[A(x), B(y)]$ then the statement usually ($V_1$ is A or $V_2$ is B) which is equivalent to usually $(V_1, V_2)$ is $H^*$ translates into the Poss-Prob granule $(V_1, V_2)$ is $m^*$ where $m^*$ is a bpa on X × Y such that $m^*(H^*) = \alpha$ and $m^*(X \times Y) = 1-\alpha$.

Similarly usually ($V_1$ is A and $V_2$ is B) translates into $(V_1, V_2)$ is $m^\perp$ where $m^\perp$ is a bpa on X × Y such that $m^\perp(H^\perp) = \alpha$ and $m^\perp(X \times Y) = 1-\alpha$ where $H^\perp$ is a fuzzy subset of X such that $H^\perp(x, y) = \text{Min}[A(x), B(y)]$.

In the above we have essentially applied this new usuality qualification operation to statements which are canonical forms from the theory of approximate reasoning, ie. V is A or $(V_1, V_2)$ is H. All the logical operations were performed on canonical statements before the usuality qualification transformed them into granules. In the next section we shall look at situations in which we combine under various logical operations structures which are of the form of Poss-Prob granules. This will enable us to translate compound statements in which the usuality qualification is more deeply embedded in the structure.

## Logical Translation Rules

Let $\perp$ be any operation definable in terms of operations on sets. Based upon the work of Yager [10] we can extend this operation to apply to Poss-Prob granules. Assume $V_1$ is $m_1$ and $V_2$ is $m_2$ are two Poss-Prob defined over the sets X and Y respectively, the rule developed by Yager states that $(V_1$ is $m_1) \perp (V_2$ is $m_2)$ translates into $(V_1, V_2)$ is $m_1 \perp m_2$ where $m = m_1 \perp m_2$ is a bpa defined on $X \times Y$ such that for all the focal elements $A_k$ of $m_1$ and $B_j$ of $m_2$

$$m(A_k \perp B_j) = m_1(A_k) * m_2(B_j).$$

Equivalently we can define m such that for any $A \subset X \times Y$

$$m(A) = \Sigma \; m_1(A_k) * m_2(B_j),$$

where the summation is taken over all $A_k, B_j$ such that $A_k \perp B_j = A$.
We note in the special case where $V_1 = V_2 = V$ defined on X then
(V is $m_1$) $\perp$ V is $m_2$ translates into V is m where m is a bpa on X defined as in the above. Dempster's rule is a special case when $\perp = \cap$.

Let us use this rule to translate some linguistic statements which involve an embedded usuality qualification.

Consider the statement "if $V_1$ is A then usually $V_2$ is B." This statement can be seen to be of the form "if $V_1$ is $m_1$ then $v_2$ is $m_2$," where $m_1$ is a bpa on X such that $m_1(A) = 1$ and $m_2$ is a bpa on Y such that $m_2(B) = \alpha$ and $m_2(Y) = 1-\alpha$.

Applying our translation rule to this situation we get, $(V_1, V_2)$ is m where $m(A_k \perp B_j) = m_1(A_k) * m_2(B_j)$ and $A_k \perp B_j = G_{kj}$. $G_{kj}$ is a fuzzy subset of $X \times Y$ such that $G_{kj}(x, y) = \text{Min} [1, 1-A_k(x) + B_j(y)]$.

In the case we are interested in with $A \perp B = D$ and $A \perp Y = X \times Y$ we get $(V_1, V_2)$ is $m^*$ where $m^*$ is a bpa on $X \times Y$ such that $m^*(D) = \alpha$ and $m^*(X \times Y) = 1-\alpha$. We note in this case $D(x, y) = \text{Min} (1, 1-A(x) + B(y))$. Parenthetically we note that this is the same translation as the statement "usually if $V_1$ is A then $V_2$ is B."

Consider next the statement, "if usually $V_1$ is A then $V_2$ is B." This can be translated into "if $V_1$ is $m_1$ then $V_2$ is $m_2$" where $m_1$ is on X such that $m_1(A) = \alpha$ and $m_1(X) = 1-\alpha$. $m_2$ is on Y such that $m_2(B) = 1$.
Using our translation rule we get $(V_1, V_2)$ is $m^*$. $m^*$ is a bpa on X such that $m^*(D) = \alpha$ and $m^*(H) = 1-\alpha$ in which $D(x, y) = \text{Min} [1, 1-A(x) + B(y)]$ and $H(x, y) = B(y)$.

Next consider the statement "if usually $V_1$ is A then usually $V_2$ is B." This can be first translated as "if $V_1$ is $m_1$ then $V_2$ is $m_2$" where $m_1$ is a bpa on X such that $m_1(A) = \alpha$ and $m_1(X) = 1-\alpha$ and $m_2$ is a bpa on Y such that $m_2(B) = \alpha$ and $m_2(X) = 1-\alpha$.

Using our logical translation rule this becomes $(V_1, V_2)$ is $m^*$, where $m^*$ is a bpa on $X \times Y$ such that $m^*(A \perp B) = \alpha^2$, $m^*(A \perp Y) = \alpha(1-\alpha)$, $m^*(X \perp B) = (1-\alpha)\alpha$ and $m^*(X \perp Y) = (1-\alpha)(1-\alpha)$. Since $A \perp B = D$, $A \perp Y = X \perp Y$ and $X \perp B = H$, then $m^*$ can be seen to be $m^*(D) = \alpha^2$, $m^*(H) = (1-\alpha)\alpha$ and $m^*(X \times Y) = 1-\alpha$.

Consider next the proposition "usually $V_1$ is A <u>and</u> usually $V_2$ is B." Formally this becomes $V_1$ is $m_1$ and $V_2$ is $m_2$ where $m_1$ is a bpa on X such that $m_1(A) = \alpha$ and $m_1(X) = 1-\alpha$. $m_2$ is a bpa on Y where $m_2(B) = \alpha$ and $m_2(Y) = 1-\alpha$. This can be seen to be equivalent to $(V_1, V_2)$ is $m^*$ where $m^*$ is on $X \times Y$ such that $m^*(D_1) = \alpha^2$, $m^*(D_2) = \alpha(1-\alpha)$, $m^*(D_3) = \alpha(1-\alpha)$ and $m^*(D_4) = (1-\alpha)^2$. In this structure $D_1(x,y) = \text{Min}(A(x), B(y))$, $D_2(x, y) = A(x)$, $D_3(x, y) = B(y)$ and $D_4(x, y) = 1$.

On the other hand "$V_1$ is $m_1$ <u>or</u> $V_2$ is $m_2$" becomes $(V_1, V_2)$ is $m^\circ$ where $m^\circ(D) = \alpha^2$ and $m^\circ(X \times Y) = 1-\alpha^2$ where $D(x, y) = \text{Max} [A(x), B(y)]$.

## Reasoning With Usual Values

In this section we shall look at the structure of some examples of reasoning with usual values. Consider the following two propositions

$P_1$: Usually [if $V_1$ is A then $V_2$ is B]
$P_2$: $V_1$ is C.

In the above we are assuming that A and C are fuzzy subsets of X, the base set of $V_1$ and B is a fuzzy of Y, the base set of $V_2$. These two pieces of data can be written as Poss-Prob granules,

$P_1$: $(V_1, V_2)$ is $m_1$ and $P_2$: $(V_1)$ is $m_2$.

In the above $m_1$ is a bpa on $X \times Y$ such that $m_1(H) = \alpha$ and $m_1(X \times Y) = 1-\alpha$ where $H(x, y) = (1-A(x)) \vee B(y)$. $m_2$ is a bpa on X such that $m_2(C) = 1$. Taking the conjunction of these two pieces of data we get "$(V_1, V_2)$ is $m_3$" where $m_3$ is a bpa on $X \times Y$ such that $m_3(E) = \alpha$ and $m_3(E_1) = 1-\alpha$ where

$$E(x, y) = H(x, y) \wedge C(x) = ((1-A(x)) \vee B(y)) \wedge C(x)$$

and $E_1(x, y) = C(x)$. Finally to get the inferred value of $V_2$, $V_2$ is $m_4$, we take the projection of $m_3$ on Y. Thus $m_4$ is a bpa on Y such that $m(F) = \alpha$ and $m(F_1) = 1-\alpha$, where $F_1 = \text{Proj}_Y E_1 \rightarrow F_1(y) = \text{Max}_x E_1(x, y) = 1$, thus $F_1 = Y$. $F = \text{Proj}_Y E$ $F(y) = \text{Max}_x [(1-A(x)) \vee B(y)) \wedge C(x)$. Thus the inferred information about $V_2$ from $P_1$ and $P_2$ is that usually $V_2$ is F.

We note that in the special case when A = C we get $F = \text{Max}_x [(A^-(x) \wedge A(x)) \vee (B(y) \wedge A(x))$. Furthermore if A is crisp then $F = B$.

Consider next the situation in which both propositions involve usual values

$P_1$: Usually (if V is A then $V_2$ is B)
$P_2$: Usually (V is C).

In this case as in the previous case $P_1$: $(V_1, V_2)$ is $m_1$ where $m_1$ is the bpa on $X \times Y$ such that $m_1(H) = \alpha$ and $m_1(X \times Y) = 1-\alpha$. However in this case

$P_2$: v is $m_2^*$

where $m_2^*$ is a bpa on X such that $m_2^*(C) = \alpha$ and $m_2^*(X) = 1-\alpha$. Taking the conjunction of these two pieces of data we get

$(V_1, V_2)$ is $m_3^*$

where $m_3^*(E) = \alpha^2$, $m_3^*(H) = \alpha(1-\alpha)$, $m_3^*(C \times Y) = \alpha(1-\alpha)$ and $m_3^*(X \times Y) = (1-\alpha)^2$.

Finally we can infer that

$V_2$ is $m_4^*$

where $m_4^*$ is a bpa on Y and its focal elements are obtained as the projection onto Y of the focal elements of $m_3^*$. We note

$\text{Proj}_Y [H] = \text{Proj}_Y [C \times Y] = \text{Proj}_Y [X \times Y] = Y$.

Since have already shown $\text{Proj}[E] = F$ we get

$m_4^*(F) = \alpha^2$ and $m_4^*(Y) = 1-\alpha^2$.

Thus in this situation we have obtained (usually)$^2$ ($V_2$ is F).

Consider the two Poss-Prob granules $V_2$ is $m_4$ and $V_2$ is $m_4^*$ where

$m_4(F) = \alpha$  $\quad m_4^*(F) = \alpha^2$
$m_4(Y) = 1-\alpha$  $\quad m_4^*(Y) = 1-\alpha^2$

which were obtained as a result of the preceding reasoning processes. Let us look at the plausibility and certainty measures associated with some arbitrary subset A of Y under each of these granules. From our definitions

$Pl(A) = \alpha \text{ Poss } [A/F] + (1-\alpha) \text{ Poss } [A/Y]$
$Pl^*(A) = \alpha^2 \text{ Poss } [A/F] + (1-\alpha^2) \text{ Poss } [A/Y]$

Since $\text{Poss }[A/Y] \geq \text{Poss }[A/F]$ and for any $\alpha \in [0,1]$, $\alpha \geq \alpha^2$ then it follows that $Pl^*(A) \geq Pl(A)$. Furthermore

$Bel(A) = \alpha \text{ Cert } [A/F] + (1-\alpha) \text{ Cert } [A/Y]$
$Bel^*(A) = \alpha^2 \text{ Cert } [A/F] + (1-\alpha^2) \text{ Cert } [A/Y]$.

Since $\text{Poss } [A^-/Y] \geq \text{Poss } [A^-/F]$ it follows that $\text{Cert } [A/F] \geq \text{Cert } [A/Y]$. Hence $Bel(A) \geq Bel^*(A)$. Thus we see that the case of a usually qualified proposition provides tighter bounds on the probability of events then that of a usually squared qualified proposition.

## Arithmetic Operations with Usual Operations

In the preceding sections we have mainly concerned ourselves with the manipulation of usual values under logic operations. In many cases we may have to perform mathematical or arithmetic operations on these usual values. In this section we develop the calculus necessary to perform these operations. The ability to handle both logical and arithmetic manipulations provides this approach with a very sophisticated mechanism for building expert systems.

Assume $V_1$ and $V_2$ are two variables taking their values in the real line R. Let $V_1$ is $m_1$ and $V_2$ is $m_2$ be two Poss-Prob granules in which $m_1$ and $m_2$ are basic probability assignments on R. Let $\{A_k\}$ be the focal elements of $m_1$ and let $\{B_j\}$ be the focal elements of $m_2$. We note that both the A's and B's are fuzzy subsets of R. Let "$V = V_1 \perp V_2$" where $\perp$ is any arithmetic operation, (addition +, subtraction -, multiplication *, division /, or exponentiation ). In [10] Yager has shown that in this situation "V is m" where m is a bpa on R such that for any $A \subset R$

$$m(A) = \Sigma\ m_1(A_i) * m_2(B_j),$$

where the summation is taken over all $A_i$ and $B_j$ such that $A_i \perp B_j = A$. In order to evaluate the above we must use fuzzy arithmetic [17]. In particular if E and F are two fuzzy numbers, fuzzy subsets of the real line, then $E \perp F = G$, where G is a fuzzy subset of R such that

$$G = \bigcup_{y,z \in R} \{E(y) \wedge F(z)/y \perp z\}.$$

Let us look at the situation for various forms of $m_1$ and $m_2$. Consider the case in which our knowledge is

$P_1$: Usually $V_1$ is A and $P_2$: $V_2$ is B

in which A and B are fuzzy subsets of R. First we see that these two pieces of data can be represented in terms of granules in the following way. $P_1$: $V_1$ is $m_1$ in which $m_1$ is a bpa on R defined by $m_1(A) = \alpha$ and $m_1(R) = 1-\alpha$. $P_2$: $V_2$ is $m_2$ where $m_2$ is also a bpa on R defined by $m_2(B) = 1$. If $V = V_1 \perp V_2$ then V is m where m is a bpa on R defined by $m(A \perp B) = \alpha$ and $m(R \perp B) = 1-\alpha$. Since for any mathematical operation, except division by $B = 0$, $R \perp B = R$ we get $m(A \perp B) = \alpha$ and $m(R) = 1-\alpha$, thus this translates to usually (V is $A \perp B$).

Consider next the situation in which both pieces of data involve usual values; $P_1$: usually $V_1$ is A and $P_2$: usually $V_2$ is B. In this case we get $V_1$ is $m_1$ and $V_2$ is $m_2$ where $m_1(A) = m_2(B) = \alpha$ and $m_1(R) = m_2(R) = 1-\alpha$. If $V = V_1 \perp V_2$ then V is $m^*$ where $m^*(A \perp B) = \alpha^2$, $m^*(A \perp R) = \alpha(1-\alpha)$, $m^*(R \perp B) = (1-\alpha)\alpha$ and $m^*(R \perp R) = (1-\alpha)^2$. However again since $A \perp R = R \perp B = R \perp R = R$ we get $m^*(A \perp B) = \alpha^2$ and $m^*(R) = 1-\alpha^2$. Thus this translates into

$$(\text{usually})^2\ (V \text{ is } A \perp B).$$

## Conclusion

We have presented a calculus for reasoning with usual valued knowledge. This system can provide a mechanism for implementing expert systems with commonsense knowledge.

345